\documentclass[10pt,twocolumn,letterpaper]{article}

\usepackage{wacv}              % To produce the CAMERA-READY version
\usepackage{graphicx}
\usepackage{amsmath}
\usepackage{amssymb}
\usepackage{booktabs}
\usepackage{subcaption}
\usepackage{amssymb}
\usepackage{dsfont}
\usepackage{colortbl}
\usepackage{xcolor}
\usepackage{tabularx}
\usepackage{multirow}

\usepackage[pagebackref,breaklinks,colorlinks]{hyperref}

\usepackage[capitalize]{cleveref}
\crefname{section}{Sec.}{Secs.}
\Crefname{section}{Section}{Sections}
\Crefname{table}{Table}{Tables}
\crefname{table}{Tab.}{Tabs.}

\def\wacvPaperID{2048} % *** Enter the WACV Paper ID here
\def\confName{WACV}
\def\confYear{2025}

\begin{document}

\title{VILLS \includegraphics[width=0.4cm]{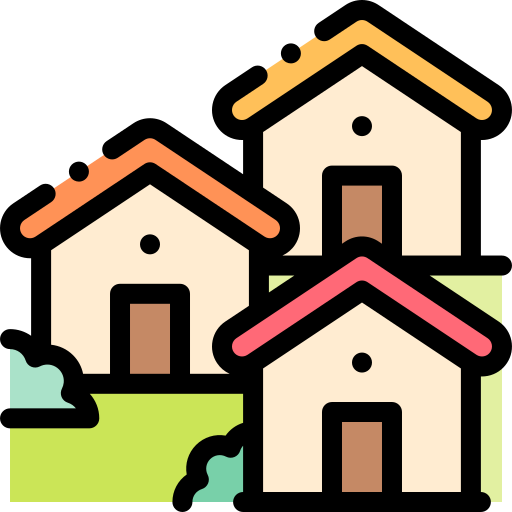}: Video-Image Learning to Learn Semantics for Person Re-Identification}

\author{Siyuan Huang, Ram Prabhakar, Yuxiang Guo, Rama Chellappa, Cheng Peng\\
Johns Hopkins University\\
{\tt\small \{shuan124, rprabha3, yguo87, rchella4, cpeng26\}@jhu.edu}
}
\maketitle

%%%%%%%%% ABSTRACT
\begin{abstract}
   Person Re-identification is a research area with significant real world applications. Despite recent progress, existing methods face challenges in robust re-identification in the wild, e.g., by focusing only on a particular modality and on unreliable patterns such as clothing. A generalized method is highly desired, but remains elusive to achieve due to issues such as the trade-off between spatial and temporal resolution and imperfect feature extraction. We propose VILLS (Video-Image Learning to Learn Semantics), a self-supervised method that jointly learns spatial and temporal features from images and videos. VILLS first designs a local semantic extraction module that adaptively extracts semantically consistent and robust spatial features. Then, VILLS designs a unified feature learning and adaptation module to represent image and video modalities in a consistent feature space. By Leveraging self-supervised, large-scale pre-training, VILLS establishes a new State-of-The-Art that significantly outperforms existing image and video-based methods. 
\end{abstract}

%%%%%%%%% BODY TEXT
\section{Introduction}
\label{sec:intro}

Person Re-identification (ReID) is a long-standing research area with many applications, e.g., in smart cities \cite{khan2024deep,behera2020person} and autonomous driving \cite{wong2020identifying,camara2020pedestrian}. The goal of ReID is to accurately match and retrieve pedestrian identities across different camera views, time periods, and locations \cite{gu2022clothes,gu2019temporal,zheng2015scalable}. Existing methods have made significant progress in improving re-identification accuracy and can generally be categorized by image and video-based methods. 

Image-based ReID methods \cite{gu2022clothes,jin2022cloth,yang2023good,chen2021person} aim to learn feature representations that can handle appearance changes by using 2D convolutions and attention mechanisms. Recent progress in self-supervised learning have further improved image-based ReID methods by allowing models to learn rich, transferable representations \cite{ericsson2022self,zhu2022pass,chen2023beyond,yuan2024hap}. These appraoches improve generalizability without relying on explicit identity labels \cite{bucci2021self,jaiswal2020survey}, which is particularly valuable as large-scale labeled datasets are rare. Video-based ReID methods \cite{wu2022cavit,cao2023event,bai2022salient} leverage temporal information with techniques such as temporal attention and long-term dependencies. Some methods also incorporate multi-modal information \cite{cao2023event} or utilize generative models \cite{xiang2022rethinking,han2023clothing} to augment training data. 

\begin{figure}[t]
    \centering
    \includegraphics[width=\linewidth]{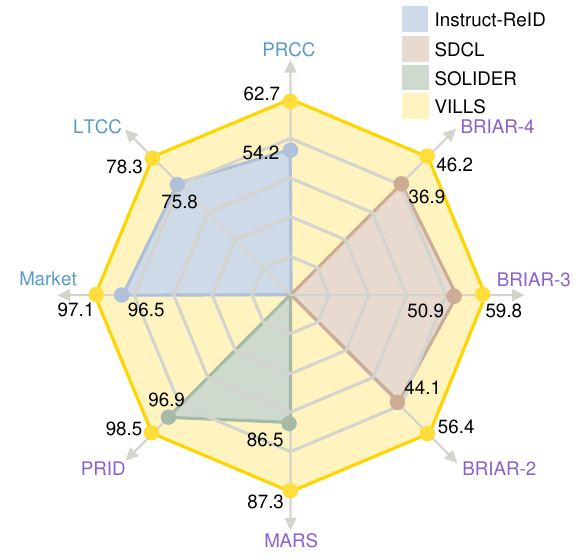}
    \caption{VILLS jointly learns spatial and temporal features from both images and videos, achieving state-of-the-art performance across \textbf{\textcolor[HTML]{559BCB}{image}} and \textbf{\textcolor[HTML]{8A60CE}{video}} ReID tasks.}
    \label{fig:radar}
\end{figure}

In principle, image-based ReID methods are good at extracting spatial features from high resolution images, while video-based methods focus on extracting temporal features with lower spatial resolution. This trade-off has been necessary due to constraints in model size, data storage, and computation. However, modern devices are capable of acquiring high-resolution videos, raising new challenges in the ReID task. Ideally, image-based ReID methods can achieve better accuracy by taking the most discriminative frame and extracting spatially fine-grained features. In practice, video-based methods are often preferred, as they are designed to aggregate multi-frame information and not affected by the choice of key frame. In this work, we explore the potential of combining the advantages of image and video-based ReID methods. 

Beyond the image-video dichotomy, ReID has to account for variations in lighting \cite{xiang2022rethinking}, pose \cite{wang2022pose}, and viewpoint \cite{zhu2020aware}. While data-driven approaches have improved general performance, nuanced scenarios, such as detecting between individuals with similar appearances or the same person with different clothing \cite{huang2023exploring}, remain difficult. As such,  extracting semantically consistent features is also of importance.

Semantically consistent features refer to identity-specific attributes that remain stable across different scenarios of the same individual and include facial structure, body proportions, motion patterns, etc. Semantically consistent features provide a more reliable basis for re-identification. By focusing on these invariant characteristics, ReID methods can achieve more robust performance. Current methods struggle to extract these features, leading to performance degradation. Through visualization (Supplemental Material), we can observe the difference between consistent and inconsistent feature attention: while inconsistent features may focus on variable elements like background, consistent features maintain attention on stable identity-specific attributes across different images or video frames of the same person.

To address these challenges, we introduce VILLS (Video-Image Learning to Learn Semantics), a self-supervised method designed to bridge the gap between image and video modalities in ReID tasks. At the core of VILLS is a Local Semantic Extraction (LSE) module, which combines a keypoint detector and an interactive segmentation model to adaptively extract semantically consistent features from images. This module allows for more nuanced and interpretable feature extraction, improving performance in challenging ReID scenarios.

VILLS extends the capabilities of LSE to videos through a Unified Feature Learning and Adaptation (UFLA) module. UFLA first uses a shared encoder to process both image and video inputs, then uses a selective resampling strategy and a feature alignment loss to align both modalities into a common feature space. This feature alignment loss addresses the inherent modality gap between images and videos by minimizing the discrepancy between the feature distributions of corresponding video and frame pairs, thereby encouraging the model to learn modality-invariant representations. Consequently, the UFLA module enables VILLS to capture complementary information from both modalities, leading to more robust and generalizable representations than previous image and video-based ReID methods. The module is trained on various large-scale unlabeled image and video datasets using a self-supervised learning formulation. By integrating LSE and UFLA, VILLS effectively extracts semantically consistent features from both images and videos, achieving significantly improved performance with a unified model. VILLS not only bridges the gap between different data modalities but also paves the way for more flexible and adaptable ReID systems applicable to real-world scenarios.

In summary, our main contributions are listed as follows:
\begin{itemize}
    \item We introduce a Local Semantic Extraction module that improves the model's ability to capture semantically consistent features across modalities. This module improves the interpretability and effectiveness of the extracted features, particularly in challenging ReID scenarios, and is seamlessly integrated into our unified framework for images and videos.
    \item We propose VILLS, a self-supervised method that unifies image and video ReID, addressing the limitations of modality-specific methods. VILLS uses a Unified Feature Learning and Adaptation module that combines shared encoding, selective resampling, and self-supervised learning to bridge the gap between images and videos while extending the capabilities of LSE to videos.
    \item Experiments across eight diverse datasets, including three distinct downstream ReID tasks that span both image and video domains, demonstrate VILLS's superior performance, with significant improvements in key metrics such as rank-1 accuracy, mAP, and TAR@0.01\% FAR. Notably, in certain tasks, VILLS outperforms existing state-of-the-art ReID methods by 9.3\%, 5.7\%, and 6.8\% on these metrics, respectively.
\end{itemize}

%-------------------------------------------------------------------------
\section{Related Work}

Image-based ReID methods have primarily focused on developing feature representations that are invariant to appearance changes. Gu et al. \cite{gu2022clothes} proposed a clothes adversarial loss to focus on discriminative features, while Yang et al. \cite{yang2023good} introduced a causality-based model to address appearance bias. Han et al. \cite{han2023clothing} explored the use of feature augmentation for expanding clothing-change data in person images. However, they are limited to processing single images and cannot leverage temporal information available in videos.

\begin{figure*}[t]
    \centering
    \includegraphics[width=\linewidth]{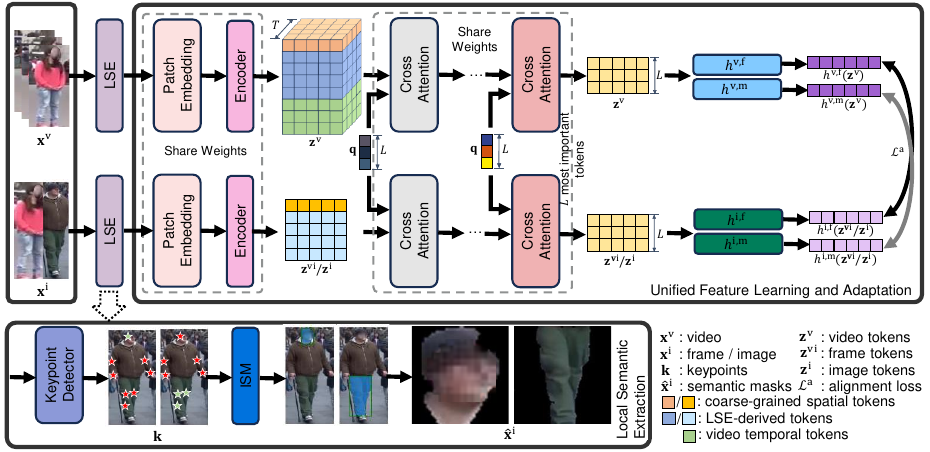}
    \caption{The pipeline of proposed VILLS. We introduce a Local Semantic Extraction module that can adaptively extract semantically consistent features from any area. Then, we introduce a Unified Feature Learning and Adaptation module to uniformly represent and jointly learn image and video semantics.}
    \label{fig:pipeline}
\end{figure*}

Video-based ReID methods aim to exploit temporal information to improve identification accuracy. Wu et al. \cite{wu2022cavit} proposed a contextual alignment model to capture spatio-temporal interaction and long-term dependencies in videos. Cao et al. \cite{cao2023event} introduced a sparse-dense complementary learning framework that integrates both sparse and dense motion information. Bai et al. \cite{bai2022salient} developed a salient-to-broad module to learn the spatio-temporal attention regions directly from videos. However, they struggle to fully leverage the fine-grained spatial details available in high-resolution videos. This limitation potentially compromises their ability to extract discriminative features that could be crucial for accurate re-identification in complex scenarios.

Recent works have begun to explore pre-training methods to learn transferable representations. Zhu et al. \cite{zhu2022pass} proposed a pre-training method that generates different level of features to offer multi-grained information. Chen et al. \cite{chen2023beyond} utilized a pre-training method to learn a general human representation from massive unlabeled human images. Yuan et al. \cite{yuan2024hap} demonstrated the effectiveness of pre-training in concentrating more on body structure information. However, these methods  struggle with the challenge of extending these features consistently to videos.

Finally, the importance of local features in ReID has been widely recognized. Zhang et al. \cite{zhang2022fine} proposed a fine-grained multi-feature fusion Network to identify and extract global and partial features. Zhang et al. \cite{zhang2022person} introduced spatial aggregation on local human parts to facilitate the strong spatial misalignment tolerance. Yan et al. \cite{yan2023clip} introduced fine-grained information excavation for mining identity clues and correspondences. However, these methods struggle with the varying importance of different spatial regions across diverse person images. Hence, they cannot provide semantically consistent features.

In summary, while existing works have made significant progress in image-based and video-based ReID separately, there remains a gap in unifying these modalities within a single framework. Additionally, the potential of self-supervised learning and semantically consistent feature extraction in multi-modal ReID has not been fully explored. Our method, VILLS, addresses these limitations by proposing a unified pre-training method that jointly learns from both image and video data, leveraging self-supervised learning and adaptive local semantic extraction.

\section{Methods}
As shown in Fig. \ref{fig:pipeline}, VILLS consists of two main components: (1) a Local Semantic Extraction Module that adaptively extracts semantically consistent features, and (2) a Unified Feature Learning and Adaptation Module that seamlessly processes both images and videos while extending LSE to videos and incorporates self-supervised learning for pre-training on large-scale unlabeled data. 

\subsection{Local Semantic Extraction Module}
To address the challenge of extracting semantically consistent and consistent features across images and videos, we propose a local semantic extraction module. This module leverages a keypoint detector and an interactive segmentation model (ISM) to adaptively focus on discriminative areas.

Specifically, for an image or video frame $\mathbf{x}^\mathrm{i}$, we use a keypoint predictor $K$ to estimate the positions of all keypoints $\mathbf{k}$, i.e.,
\begin{equation}
    \mathbf{k} = K(\mathbf{x}^\mathrm{i}) = [\mathbf{k}_1, ..., \mathbf{k}_n],
\label{eq:predictor}
\end{equation}
where $n$ is the number of keypoints. The predictor automatically determines the meaning of each keypoint. For example, if the predictor is trained on COCO \cite{lin2014microsoft}, $\mathbf{k}_1$ represents nose, $\mathbf{k}_2$ represents left eye, and so on.

ISMs \cite{kirillov2023segment,zou2023segment,ravi2024sam} have demonstrated powerful and flexible segmentation capabilities, allowing for image segmentation at any area and granularity given appropriate prompts. We leverage an ISM to extract fine-grained spatial features. Using the detected keypoints, we construct prompts for the ISM. For an given area $a$, we identify the keypoints within it as positive and the others as negative, creating an indicator vector
\begin{equation}
    \mathbf{v}^a = \mathds{1}_a (\mathbf{k}) = [\mathbf{v}^a_1, ..., \mathbf{v}^a_n],
\end{equation}
where $\mathds{1}_{\mathrm{cond}}$ is the indicator function. We then input $\mathbf{x}^\mathrm{i}$, $\mathbf{k}$, and $\mathbf{v}^a$ into the ISM $M$ to obtain the spatial features corresponding to area $a$, i.e.,
\begin{equation}
    \mathbf{x}^{\mathrm{i}, a} = M (\mathbf{x}^\mathrm{i}, \mathbf{k}, \mathbf{v}^a).
    \label{eq:ism}
\end{equation}
This method enables extraction of semantically consistent features from any area. For example, to obtain head features, we can set the nose, eye, and ear keypoints as positive and the others as negative.

However, keypoints far from the target position can decrease feature accuracy when used as prompts. For instance, given a torso area, distant keypoints result in background noise and thighs are also included. To address this issue, we introduce a filtering mechanism. The keypoint predictor provides a confidence score $\mathbf{s} = [\mathbf{s}_1, ..., \mathbf{s}_n]$ for each keypoint. We filter $\mathbf{v}^a$ using a threshold $\tau^a$ on $\mathbf{s}$, i.e.,
\begin{equation}
    \mathbf{v}^a = \mathds{1}_{\mathbf{s} > \tau^a \cap a} (\mathbf{k}),
\end{equation}
and apply it to Eq. \ref{eq:ism}. The final set of semantically consistent features is defined as
\begin{equation}
    \hat{\mathbf{x}}^{\mathrm{i}} = \{ \mathbf{x}^{\mathrm{i}, a} | a \in \mathcal{A} \},
\end{equation}
where $\mathcal{A}$ is the set of areas.

\subsection{Unified Feature Learning and Adaptation Module}
To bridge the gap between image and video modalities and extend LSE to videos, we propose a Unified Feature Learning and Adaptation (UFLA) module. This module leverages a shared encoder and selective resampling to represent features of both modalities in a common feature space, then integrated with self-supervised learning to learn from large-scale unlabeld datasets.

Specifically, given an original video $\mathbf{x}^\mathrm{v}$ and an original image $\mathbf{x}^\mathrm{i}$, the shared encoder $E$ converts the inputs into video tokens $\mathbf{z}^\mathrm{v}$ and image tokens $\mathbf{z}^\mathrm{i}$, i.e.,
\begin{equation}
    \mathbf{z}^\mathrm{v} = E(\mathbf{x}^\mathrm{v}), \mathbf{z}^\mathrm{i} = E(\mathbf{x}^\mathrm{i}),
\label{eq:backbone}
\end{equation}
where $\mathbf{z}^\mathrm{v} \in \mathbb{R}^{T \times L^{\mathrm{v}} \times d}$ and $\mathbf{z}^\mathrm{i} \in \mathbb{R}^{L^{\mathrm{i}} \times d}$. Here, $T$ is the number of frames, $L^{\mathrm{v}}$ and $L^{\mathrm{i}}$ represent the number of video and image tokens, respectively, and $d$ is the embedding dimension. The video tokens $\mathbf{z}^\mathrm{v}$ contains three types of semantics: (1) coarse-grained spatial semantics, representing global features of each frame; (2) LSE-derived semantics, capturing local semantically consistent features from the LSE module; and (3) temporal semantics, encoding motion patterns of the video. In contrast, the image tokens $\mathbf{z}^\mathrm{i}$ contain only the first two types of semantics, lacking temporal information.

At this moment, $\mathbf{z}^\mathrm{v}$ and $\mathbf{z}^\mathrm{i}$ are not aligned, which poses a significant challenge for joint learning. This misalignment can lead to suboptimal feature representations, reducing the model's ability to generalize across image and video domains. Moreover, it complicates the design of loss functions.

To address this issue and align the feature representations of both modalities in a common feature space, we first sample one frame from each video to construct frame tokens $\mathbf{z}^\mathrm{vi}$. These frame tokens have the same structure as the image tokens $\mathbf{z}^\mathrm{i}$. We then introduce a selective resampling strategy \cite{jaegle2021perceiver,alayrac2022flamingo,team2023gemini} to represent them using a resampler. The resampler $R$ uses cross-attention to select the most important $L$ tokens for each type of input, i.e.,
\begin{equation}
    \mathbf{z}^\mathrm{v} \leftarrow R(\mathbf{z}^\mathrm{v}; \mathbf{q}), \mathbf{z}^\mathrm{vi} \leftarrow R(\mathbf{z}^\mathrm{vi}; \mathbf{q}), \mathbf{z}^\mathrm{i} \leftarrow R(\mathbf{z}^\mathrm{i}; \mathbf{q}),
\label{eq:resampler}
\end{equation}
where $\mathbf{q}$ is the learnable query. For video tokens, the resampler first flattens the temporal and spatial dimensions, resulting in $\mathbf{z}^\mathrm{v} \in \mathbb{R}^{(T \cdot L^{\mathrm{v}}) \times d}$, then selects the most important $L$ tokens. For frame and image tokens, the resampler directly selects $L$ tokens. This method ensures that the resampler can adaptively focus on the most relevant information from both modalities, regardless of their original number of tokens, effectively extending the LSE to the video domain.

The resulting uniform representation allows subsequent computations to be performed directly using the same model and loss function, regardless of the input modality. Hence, VILLS can better leverage the complementary information present in images and videos, leading to more robust and generalizable representations for ReID tasks.

Then, we use self-supervised learning to learn $\mathbf{z}^\mathrm{v}$, $\mathbf{z}^\mathrm{vi}$, and $\mathbf{z}^\mathrm{i}$ from large-scale unlabeled datasets. DINO series methods \cite{caron2021emerging,oquab2023dinov2} have demonstrated effective self-supervised semantic learning capabilities, which we leverage for jointly learning image and video features. 

Specifically, the shared encoder $E$ in (\ref{eq:backbone}) consists of two encoders with identical structures. The teacher encoder's input is the original video $\mathbf{x}^\mathrm{v}$ and the original image $\mathbf{x}^\mathrm{i}$, while the student encoder's input includes $\mathbf{x}^\mathrm{v}$, $\mathbf{x}^\mathrm{i}$, and their LSE features $\hat{\mathbf{x}}^\mathrm{v}$ and $\hat{\mathbf{x}}^\mathrm{i}$. Hence, the video tokens $\mathbf{z}^\mathrm{v}$ and image tokens $\mathbf{z}^\mathrm{i}$ in (\ref{eq:backbone}) become
\[
\begin{bmatrix}
    \mathbf{z}^\mathrm{v} \\
    \mathbf{z}^\mathrm{i}
\end{bmatrix}
=
\begin{bmatrix}
    \mathbf{z}^\mathrm{v, t}, \mathbf{z}^\mathrm{v, s} \\
    \mathbf{z}^\mathrm{i, t}, \mathbf{z}^\mathrm{i, s}
\end{bmatrix}
= [\mathbf{z}^\mathrm{t}, \mathbf{z}^\mathrm{s}]^\intercal,
\]
where $\mathrm{t}$ and $\mathrm{s}$ denote teacher and student, respectively. 

We apply the resampling in (\ref{eq:resampler}) to $\mathbf{z}^\mathrm{v}$, $\mathbf{z}^\mathrm{vi}$, and $\mathbf{z}^\mathrm{i}$ to obtain the resampled tokens. Then, we use various self-supervised losses to compute the feature distributions. The feature head 
\[
h^\mathrm{f}
=
\begin{bmatrix}
    h^\mathrm{v, f} \\
    h^\mathrm{i, f}
\end{bmatrix}
=
\begin{bmatrix}
    h^\mathrm{v, f, t}, h^\mathrm{v, f, s} \\
    h^\mathrm{i, f, t}, h^\mathrm{i, f, s}
\end{bmatrix}
= [h^\mathrm{f, t}, h^\mathrm{f, s}]^\intercal,
\]
calculates the feature distributions, and the feature loss $\mathcal{L}^{\mathrm{f}}$ measures the distribution difference between the teacher and student, aiming to make the student's distribution consistent with the teacher's, i.e.,
\begin{equation}
    \mathcal{L}^{\mathrm{f}} = \mathcal{H}(h^\mathrm{f, t}(\mathbf{z}^\mathrm{t}), h^\mathrm{f, s}(\mathbf{z}^\mathrm{s})),
\end{equation}
where $\mathcal{H}(\mathbf{a}, \mathbf{b}) = - \sum_j \mathbf{a}_j \log \mathbf{b}_j$ represents entropy, and $\mathbf{a}$ and $\mathbf{b}$ are the two distributions.

The masking head $h^\mathrm{m}$ calculates the masked feature distributions, and the masking loss $\mathcal{L}^{\mathrm{m}}$ measures the distribution difference between the unmasked teacher and the masked student, encouraging the student to infer the masked tokens based on unmaksed tokens, i.e.,
\begin{equation}
    \mathcal{L}^{\mathrm{m}} = \mathcal{H}(h^\mathrm{m, t}(\mathbf{z}^\mathrm{t}), h^\mathrm{m, s}(\mathrm{mask}(\mathbf{z}^\mathrm{s}))),
\end{equation}
where $\mathrm{mask} (\cdot)$ refers to the masking function.

The regularization loss $\mathcal{L}^{\mathrm{r}}$ is used to regularize the tokens, improving the generalization and stability of the learned semantics. We use KoLeo regularization \cite{sablayrolles2018spreading,oquab2023dinov2} $k$ to smooth the student tokens, i.e.,
\begin{equation}
    \mathcal{L}^{\mathrm{r}} = k(\mathbf{z}^\mathrm{s}).
\end{equation}

Finally, we introduce an alignment loss $\mathcal{L}^{\mathrm{a}}$ to align video and frame tokens, ensuring the model to learn consistent representations across different modalities derived from the same video source. Inspired by CLIP \cite{radford2021learning}, we design a pair alignment loss to minimize the distance between video distribution and frame distribution from the same video, i.e.,
\begin{equation}
    \mathcal{L}^{\mathrm{a}} = \frac{1}{2} (\mathrm{CE}(h^\mathrm{v}(\mathbf{z}^\mathrm{v}, \mathbf{y})) + \mathrm{CE}(h^\mathrm{i}(\mathbf{z}^\mathrm{vi}, \mathbf{y})),
\end{equation}
where $h^\mathrm{v}$ and $h^\mathrm{i}$ are all heads for videos and images respectively, $\mathbf{y}$ represents the video indices, and $\mathrm{CE}$ denotes the cross entropy loss. 

The overall loss for self-supervised learning is used to 
\begin{equation}
    \mathcal{L} = \lambda_1 \mathcal{L}^{\mathrm{f}} + \lambda_2 \mathcal{L}^{\mathrm{m}} + \lambda_3 \mathcal{L}^{\mathrm{r}} + \lambda_4 \mathcal{L}^{\mathrm{a}},
\end{equation}
where $\lambda$ are balancing parameters. The loss function only updates the student, and the teacher is updated from the student weights using an exponential moving average \cite{he2020momentum}. This strategy allows the model to effectively learn semantics, resulting in the teacher having more general, robust, and smooth representations \cite{caron2021emerging}. We now completed the entire process for VILLS.

\section{Experiments}

\subsection{Datasets and Evaluation Metrics}
For pre-training, we utilize two large-scale datasets. For images, we use LUPerson \cite{fu2021unsupervised}, a dataset spcificially designed for ReID tasks, comprising 4,180,243 person images. For videos, given the absense of publicly available large-scale ReID video datasets, we created a custom dataset based on Kinetics-700 (K-700) \cite{carreira2019short}. K-700 is a comprehensive action recognition dataset containing 700 actions across approximately 650,000 videos. To address the challenge of background noise and multiple intersecting individuals in K-700, we applied a multi-object tracking method \cite{chen2024delving} with a confidence threshold of 0.8. This preprocessing step resulted in a high-quality ReID video dataset of 67,956 videos, effectively preserving temporal information while minimizing noise.

We evaluate VILLS on eight diverse datasets spanning image and video ReID tasks: PRCC \cite{yang2019person}, LTCC \cite{qian2020long}, PRID2011 \cite{hirzer2011person}, MARS \cite{zheng2016mars}, BRIAR-2 \cite{cornett2023expanding}, BRIAR-3 \cite{cornett2023expanding}, BRIAR-4 \cite{cornett2023expanding}, and Market1501 \cite{zheng2015scalable}. Among these, PRCC, LTCC, and Market1501 are image-only datasets, while PRID2011 and MARS are video-only datasets. BRIAR-2, BRIAR-3, and BRIAR-4 represent image-video mix datasets.

\begin{table*}[t]
    \centering
    \begin{subtable}[t]{0.592\textwidth}
        \centering
        \resizebox{\linewidth}{!}{
        \begin{tabular}{lcccccc}
        \hline
         & \multicolumn{2}{c}{PRCC} & \multicolumn{2}{c}{LTCC} & \multicolumn{2}{c}{Market1501} \\ \cline{2-7} 
        Method & mAP & R-1 & mAP & R-1 & mAP & R-1 \\ \hline
        RCSANet \cite{huang2021clothing} & 48.6 & 50.2 & / & / & / & / \\
        SFA \cite{li2021simple} & 47.8 & 49.6 & 33.6 & 61.7 & / & / \\
        FSAM \cite{hong2021fine} & / & 54.5 & 35.4 & 73.2 & / & / \\
        CAL \cite{gu2022clothes} & 54.4 & 54.4 & 39.4 & 73.4 & / & / \\
        AIM \cite{yang2023good} & \underline{58.3} & 57.9 & 41.1 & 76.3 & / & / \\
        CCFA \cite{han2023clothing} & \textbf{58.4} & \underline{61.2} & \underline{42.5} & 75.8 & / & / \\
        SCSN \cite{chen2020salience} & / & / & / & / & 88.5 & 95.7 \\
        ISP \cite{zhu2020identity} & / & / & / & / & 88.6 & 95.3 \\
        RGA-SC \cite{zhang2020relation} & / & / & / & / & 88.4 & 96.1 \\
        MoCoV2 \cite{chen2020improved} & / & / & / & / & 91.0 & 96.4 \\
        TransReID \cite{he2021transreid} & / & / & / & / & 88.9 & 95.2 \\
        PHA \cite{zhang2023pha} & / & / & / & / & 90.2 & 96.1 \\ \hline
        PASS \cite{zhu2022pass} & 53.4 & 52.4 & 38.2 & 74.8 & 92.3 & 96.8 \\
        SOLIDER$^\dag$ \cite{chen2023beyond} & 49.9 & 50.1 & 34.9 & 72.4 & \textbf{93.9} & \underline{96.9} \\
        UniHCP \cite{ci2023unihcp} & / & / & / & / & 90.3 & / \\
        HAP \cite{yuan2024hap} & 45.9 & 45.4 & 35.0 & 71.2 & 91.7 & 96.1 \\
        Instruct-ReID$^*$ \cite{he2024instruct} & 52.3 & 54.2 & \textbf{52.0} & 75.8 & \underline{93.5} & 96.5 \\ \hline
        \textcolor[HTML]{0055A4}{VILLS (Image only) (ours)} & \textcolor[HTML]{0055A4}{55.0} & \textcolor[HTML]{0055A4}{58.4} & \textcolor[HTML]{0055A4}{39.0} & \textcolor[HTML]{0055A4}{\underline{77.7}} & \textcolor[HTML]{0055A4}{92.9} & \textcolor[HTML]{0055A4}{96.8} \\
        \textcolor[HTML]{0055A4}{VILLS (ours)} & \textcolor[HTML]{0055A4}{58.0} & \textcolor[HTML]{0055A4}{\textbf{62.7}} & \textcolor[HTML]{0055A4}{39.3} & \textcolor[HTML]{0055A4}{\textbf{78.3}} & \textcolor[HTML]{0055A4}{92.9} & \textcolor[HTML]{0055A4}{\textbf{97.1}} \\ \hline
        \end{tabular}
        }
        \caption{Image-based ReID}
    \end{subtable}
    \hfill
    \begin{minipage}[t]{0.403\textwidth}
        \centering
        \vspace{-4.3cm}
        % First subtable on the right
        \begin{subtable}[t]{\textwidth}
            \centering
            \resizebox{\linewidth}{!}{
            \begin{tabular}{lcccc}
            \hline
             & \multicolumn{2}{c}{PRID2011} & \multicolumn{2}{c}{MARS} \\ \cline{2-5} 
            Method & mAP & R-1 & mAP & R-1 \\ \hline
            MG-RAFA \cite{zhang2020multi} & / & 95.9 & 85.9 & 88.8 \\
            GRL \cite{liu2021watching} & 93.2 & 87.6 & 82.8 & 88.7 \\
            STRF \cite{aich2021spatio} & / & / & 86.1 & / \\
            STMN \cite{eom2021video} & 94.0 & 91.0 & 83.4 & 89.0 \\
            PSTA \cite{wang2021pyramid} & 94.7 & 93.3 & 85.1 & 89.9 \\
            SINet \cite{bai2022salient} & / & 96.5 & 86.2 & 91.0 \\
            GI-ReID \cite{jin2022cloth} & / & / & 80.4 & / \\
            CAViT \cite{wu2022cavit} & \underline{97.3} & 95.5 & \underline{87.2} & \underline{90.8} \\
            SDCL \cite{cao2023event} & 96.9 & \underline{96.5} & 86.5 & \textbf{91.1} \\ \hline
            \textcolor[HTML]{0055A4}{VILLS (ours)} & \textcolor[HTML]{0055A4}{\textbf{98.5}} & \textcolor[HTML]{0055A4}{\textbf{97.8}} & \textcolor[HTML]{0055A4}{\textbf{87.3}} & \textcolor[HTML]{0055A4}{90.7} \\ \hline
            \end{tabular}
            }
            \caption{Video-based ReID}
        \end{subtable}
        
        % Vertical space to balance the height
        \vspace{0.73cm}

        % Second subtable on the right (below the first)
        \begin{subtable}[t]{\textwidth}
            \centering
            \resizebox{\linewidth}{!}{
            \begin{tabular}{lccccc}
            \hline
             &  & \multicolumn{2}{c}{PRID2011} & \multicolumn{2}{c}{MARS} \\ \cline{3-6} 
            Method & Query & mAP & R-1 & mAP & R-1 \\ \hline
            \multirow{2}{*}{CAViT \cite{wu2022cavit}} & Image & 92.2 & 88.8 & 78.6 & 84.7 \\
             & Video & 97.3 & 95.5 & 87.2 & \textbf{90.8} \\ \hline
            % \rowcolor[HTML]{0055A4}
            % \rowcolors{1}{[HTML]{0055A4}}{} % Apply color from this row onwards
            \multirow{2}{*}{\textcolor[HTML]{0055A4}{VILLS (ours)}} & \textcolor[HTML]{0055A4}{Image} & \textcolor[HTML]{0055A4}{97.8} & \textcolor[HTML]{0055A4}{96.6} & \textcolor[HTML]{0055A4}{82.9} & \textcolor[HTML]{0055A4}{87.3} \\
            % \rowcolor[HTML]{0055A4}
             & \textcolor[HTML]{0055A4}{Video} & \textcolor[HTML]{0055A4}{\textbf{98.5}} & \textcolor[HTML]{0055A4}{\textbf{97.8}} & \textcolor[HTML]{0055A4}{\textbf{87.3}} & \textcolor[HTML]{0055A4}{90.7} \\ \hline
             % \rowcolors{4}{}{} % Stop coloring
            \end{tabular}
            }
            \caption{Video-based ReID with Different Queries}
        \end{subtable}
    \end{minipage}
    \vfill
    \begin{subtable}[t]{\textwidth}
        % \vspace{0.5cm}
        \centering
        \resizebox{\linewidth}{!}{
        \begin{tabular}{lcccccccccccc}
        \hline
         & \multicolumn{4}{c}{BRIAR-2} & \multicolumn{4}{c}{BRIAR-3} & \multicolumn{4}{c}{BRIAR-4} \\ \cline{2-13} 
        Method & R-1 & R-20 & T@0.01\%F & T@1\%F & R-1 & \multicolumn{1}{c}{R-20} & T@0.01\%F & \multicolumn{1}{c}{T@0.1\%F} & R-1 & \multicolumn{1}{c}{R-20} & T@0.01\%F & \multicolumn{1}{c}{T@0.1\%F} \\ \hline
        TranReID \cite{he2021transreid} & 25.0 & 70.3 & 5.1 & 50.0 & / & / & / & / & / & / & / & / \\
        PFD \cite{wang2022pose} & 32.9 & 75.7 & 8.5 & 48.0 & / & / & / & / & / & / & / & / \\
        DC-Former \cite{wang2022pose} & 28.0 & 72.9 & 7.5 & 49.4 & / & / & / & / & / & / & / & / \\
        FarSight \cite{liu2023farsight} & / & 72.9 & / & 54.0 & / & / & / & / & / & / & / & / \\
        shARc \cite{zhu2023sharc} & 41.1 & 83.0 & / & / & / & / & / & / & / & / & / & / \\
        PSTA \cite{wang2021pyramid} & / & / & / & / & 27.8 & / & / & 21.5 & / & / & / & / \\
        GaitGL \cite{lin2021gait} & / & / & / & / & 12.6 & / & / & 6.4 & / & / & / & / \\
        MViTv2 \cite{li2022mvitv2} & / & / & / & / & 11.8 & / & / & 8.4 & / & / & / & / \\
        ABNet \cite{azad2024activity} & / & / & / & / & 34.4 & / & / & 26.4 & / & / & / & / \\
        CAL \cite{gu2022clothes} & 34.9 & 71.2 & / & 51.9 & 30.6 & 74.9 & 5.1 & 25.4 & 16.9 & 53.9 & 3.9 & 12.7 \\
        BRIARNet \cite{huang2023whole} & 32.8 & 75.1 & 5.4 & 54.1 & 30.6 & 80.2 & 0 & 5.0 & 15.7 & 55.5 & 0 & 3.7 \\ \hline
        HAP \cite{yuan2024hap} & / & / & / & / & / & / & / & / & 28.5 & 76.9 & 5.1 & 17.9 \\
        PASS \cite{zhu2022pass} & 44.3 & 88.6 & 14.9 & 66.3 & 45.0 & 95.1 & 8.5 & 25.3 & 35.9 & 85.7 & 9.8 & 28.1 \\
        SOLIDER$^\dag$ \cite{chen2023beyond} & 44.1 & 86.8 & 13.5 & 69.5 & 50.9 & \underline{95.6} & 9.4 & \underline{31.8} & 36.9 & 86.4 & 11.3 & 31.8 \\ \hline
        \textcolor[HTML]{0055A4}{VILLS (Image only) (ours)} & \textcolor[HTML]{0055A4}{\underline{51.2}} & \textcolor[HTML]{0055A4}{\underline{89.1}} & \textcolor[HTML]{0055A4}{\underline{21.3}} & \textcolor[HTML]{0055A4}{\underline{72.0}} & \textcolor[HTML]{0055A4}{\underline{51.4}} & \textcolor[HTML]{0055A4}{94.3} & \textcolor[HTML]{0055A4}{\underline{12.3}} & \textcolor[HTML]{0055A4}{30.4} & \textcolor[HTML]{0055A4}{\underline{40.9}} & \textcolor[HTML]{0055A4}{\underline{86.7}} & \textcolor[HTML]{0055A4}{\underline{13.5}} & \textcolor[HTML]{0055A4}{\underline{34.3}} \\
        \textcolor[HTML]{0055A4}{VILLS (ours)} & \textcolor[HTML]{0055A4}{\textbf{56.4}} & \textcolor[HTML]{0055A4}{\textbf{90.8}} & \textcolor[HTML]{0055A4}{\textbf{24.7}} & \textcolor[HTML]{0055A4}{\textbf{75.7}} & \textcolor[HTML]{0055A4}{\textbf{59.8}} & \textcolor[HTML]{0055A4}{\textbf{96.7}} & \textcolor[HTML]{0055A4}{\textbf{17.3}} & \textcolor[HTML]{0055A4}{\textbf{39.5}} & \textcolor[HTML]{0055A4}{\textbf{46.2}} & \textcolor[HTML]{0055A4}{\textbf{89.6}} & \textcolor[HTML]{0055A4}{\textbf{18.1}} & \textcolor[HTML]{0055A4}{\textbf{40.4}} \\ \hline
        \end{tabular}
        }
        \caption{Image-Video Mix ReID}
    \end{subtable}
    \caption{Main table: Evaluations on different ReID tasks. Top two results are highlighted in \textbf{bold} and \underline{underlined}. $\dag$ results used better backbone. * results used multiple pre-training datasets. In all tables, group 1 methods have specific design for their task. In table (a) and (d), group 2 are image pre-training methods.}
    \label{tab:main_tab}
    \vspace{-0.4cm}
\end{table*}

For image-based and video-based ReID, we report rank-1 accuracy and mean average precision (mAP). For image-video mix ReID, we also report rank-$k$ accuracy and true acceptance rate (TAR) at a $x\%$ false acceptance rate (FAR) to evaluate the model's ability in practical applications.

\subsection{Main Results}
\subsubsection{Image-based ReID}
Table \ref{tab:main_tab}(a) compares VILLS with existing methods on image-only datasets. The first group includes methods specifically designed for these datasets, while the second group includes pre-training methods. Since other pre-training methods only use image pre-training datasets, to ensure a fair comparison, we report results using the same image pre-training dataset and using both image and video pre-training datasets.

Compared to group 1 methods, VILLS achieves state-of-the-art (SOTA) results on several metrics. Our rank-1 accuracy is 1.5\% higher than CCFA \cite{han2023clothing} on PRCC and 2.5\% higher on LTCC. On Market-1501, our mAP is 1.9\% higher than MoCoV2 \cite{chen2020improved}. These results demonstrate the effectiveness of our method. When compared to other pre-training methods, even with image-only pre-training, VILLS still achieves SOTA results on many metrics. Our mAP is 2.7\% higher than Instruct-ReID \cite{he2024instruct} on PRCC, and our rank-1 accuracy is 1.9\% higher on LTCC. These results highlight the effectiveness of VILLS's components. When incorporating videos in pre-training, the performance advantages become even more obvious. We achieve the highest rank-1 accuracy across all datasets: 62.7\% on PRCC, 78.3\% on LTCC, and 97.1\% on Market-1501. While Instruct-ReID shows impressive mAP on LTCC, it exhibits inconsistent results across other datasets, highlighting a limitation in Instruct-ReID's capabilities in image-based ReID. Notably, VILLS achieves a rank-1 accuracy 8.5\% higher than Instruct-ReID on PRCC. These experiments validate the effectiveness of our method in image-based ReID. 

% \vspace{-0.7cm}

\subsubsection{Video-based ReID}
Table \ref{tab:main_tab}(b) compares VILLS with other methods on video-only datasets. All compared methods are specifically designed for these datasets. VILLS demonstrates SOTA performance, with our rank-1 accuracy 1.3\% higher than SDCL \cite{cao2023event} on PRID2011 and our mAP 0.8\% higher on MARS. Notably, VILLS is the first ReID pre-training method that can naturally process videos. These results not only demonstrate the effectiveness of our method but also illustrate that VILLS successfully learns video semantics.

\subsubsection{Image-Video Mix ReID}
We evaluate this scenario in two contexts: in-lab and in-the-wild. For in-lab evaluation, we create image-video mix datasets by extracting single frames from each video in the query set of video-only datasets, while maintaining the video gallery. Table \ref{tab:main_tab}(c) compares VILLS with the SOTA video-based method, CAViT \cite{wu2022cavit}. VILLS consistently outperforms CAViT, surpassing it by 2.3\% in rank-1 accuracy for video queries on PRID2011 and by 4.3\% in mAP for image queries on MARS. Moreover, VILLS demonstrates greater robustness across modalities. While CAViT's performance drops by 5.1\% in mAP on PRID2011 when switching from video to image queries, VILLS only experiences a 0.7\% decrease. This underscores our method's effectiveness in simultaneously addressing both temporal and spatial information, unlike video-based ReID methods that may overly focus on temporal aspects at the expense of spatial information.

Table \ref{tab:main_tab}(d) shows results for image-video mix datasets in-the-wild. The first group includes methods specifically designed for these datasets, while the second group includes pre-training methods. As no video-based ReID pre-training methods exist yet, we treat each video frame independently for group 2 methods and our image-only version. Even our image-only version achieves SOTA results across both groups on most metrics. Our rank-1 accuracy is 6.9\% higher than PASS \cite{zhu2022pass} on BRIAR-2. On BRIAR-3, our TAR@0.01\%FAR is 2.9\% higher than SOLIDER \cite{chen2023beyond}, making VILLS the first method to reach double digits on this metric. Our TAR@0.1\%FAR is 2.5\% higher than SOLIDER on BRIAR-4. These results strongly demonstrate the effectiveness of VILLS's components. When incorporating video pre-training, the performance advantages become even more pronounced, with VILLS achieving the highest results on every metric across all datasets. Given that VILLS is the only pre-training method capable of naturally capturing both spatial and temporal features, we believe it is particularly well-suited for image-video mix ReID tasks.

To summarize, the consistent improvements across both image and video modalities highlight the effectiveness of VILLS's unified framework in leveraging complementary information from both domains. VILLS demonstrates strong generalization capabilities across image-based ReID, video-based ReID, and image-video mix ReID tasks. These comprehensive results underscore VILLS's SOTA performance and its ability to unify image and video ReID within a single, effective framework.

\subsection{Ablation Studies}
\subsubsection{Significance of Components}
We evaluate the impact of each pre-training component on different ReID tasks. The vanilla VILLS serves as our baseline, lacking both the Local Semantic Extraction (LSE) module and the Unified Feature Learning and Adaptation (UFLA) module.

\begin{table}[t]
    \centering
    \begin{subtable}[t]{0.4\textwidth}
        \centering
        \resizebox{\linewidth}{!}{
        \begin{tabular}{lcccc}
        \hline
         & \multicolumn{2}{c}{PRCC} & \multicolumn{2}{c}{LTCC} \\ \cline{2-5} 
         & mAP & R-1 & mAP & R-1 \\ \hline
        Vanilla VILLS & 47.2 & 48.1 & 35.2 & 72.4 \\ \hline
        $|\mathcal{A}|$ = 1 & 47.4 & 48.7 & 35.5 & 73.6 \\
        $|\mathcal{A}|$ = 2 & 47.7 & 49.0 & 35.9 & 74.2 \\
        $|\mathcal{A}|$ = 3 & \textbf{48.9} & \textbf{49.7} & \textbf{36.3} & \textbf{74.2} \\ \hline
        $M$ is parsing & 54.4 & 55.7 & 37.0 & 74.2 \\
        $M$ is ISM & \textbf{54.5} & \textbf{57.0} & \textbf{37.5} & \textbf{76.1} \\ \hline
        \end{tabular}
        }
        \caption{Local Semantic Extraction Module}
    \end{subtable}
    \hfill
    \begin{subtable}[t]{0.48\textwidth}
        \centering
        \resizebox{\linewidth}{!}{
        \begin{tabular}{lcccc}
        \hline
         & \multicolumn{2}{c}{PRID2011} & \multicolumn{2}{c}{BRIAR-4} \\ \cline{2-5} 
         & mAP & R-1 & T@0.1\%F & R-1 \\ \hline
        VILLS (image only) & 95.2 & 93.3 & 23.8 & 30.8 \\
        + shared encoder & 95.5 & 93.3 & 24.8 & 32.8 \\
        + perceiver & \textbf{98.5} & \textbf{97.8} & \textbf{28.9} & \textbf{37.1} \\ \hline
        \end{tabular}
        }
        \caption{Unified Feature Learning and Adaptation Module}
    \end{subtable}
    \caption{Significance of Components.}
    \label{tab:ab}
\end{table}

First, we evaluate the LSE module on image-based ReID (Table \ref{tab:ab}(a)). The 1st group demonstrates the impact of varying the number of areas in LSE when applied to vanilla VILLS. Performance improves with a larger number of areas, peaking at three areas. Compared to vanilla VILLS, our mAP improves by 1.7\% on PRCC, and our rank-1 accuracy increases by 1.8\% on LTCC, highlighting the LSE module's effectiveness. Based on these results, we set $\mathcal{A} = 3$ for LSE in VILLS.

Given that parsing models can also provide segmentation for human areas, we compare the use of an interactive segmentation model (ISM) to a parsing model in LSE. Group 2 in Table \ref{tab:ab}(a) shows this comparison. The ISM-based LSE outperforms the parsing model, with a 1.3\% higher rank-1 accuracy on PRCC and 1.9\% higher on LTCC. This underscores the ISM's effectiveness. The key difference between ISM and parsing models is that ISM can leverage prompts to focus on more specific areas than a parsing model that focuses solely on pre-defined areas by parsing categories, leading to semantically consistent features. A visual comparison (Supplemental Material) validates this point, showing that our method focuses more on semantically consistent areas (e.g., arms and legs) while ignoring unrelated regions, unlike non-ISM methods.

Next, we evaluate the UFLA module on video-based and image-video mix ReID tasks (Table \ref{tab:ab}(b)). We add components in the UFLA module to the image version of VILLS. Applying the shared encoder improves performance across the image version of VILLS and ReID tasks. With the shared encoder, our mAP increases by 2.0\% over vanilla VILLS on PRID2011, and our rank-1 accuracy improves by 2.0\% over the image version of VILLS on BRIAR-4. Adding the perceiver further improves performance, with our rank-1 accuracy increasing by 4.5\% over the image version of VILLS on PRID2011, and our TAR@0.1\%FAR improving by 5.1\% on BRIAR-4. These results demonstrate the effectiveness of each component in the UFLA module. In summary, these ablation studies confirm the significant contributions of both the LSE and UFLA modules in VILLS.

\subsubsection{Losses and Balancing Parameters}
We conduct a comprehensive study on the impact of varying $\lambda$ parameters for self-supervised losses in the UFLA module on image-based ReID and video-based ReID ( Table \ref{tab:bal}). 

In Group 1, we examine the ratio of masking loss to feature loss by keeping $\lambda_1$ fixed and modifying $\lambda_2$. VILLS achieves optimal performance on both ReID tasks when $\lambda_2 = 1.0$. Performance decreases when $\lambda_2 = 2.0$, indicating that excessive masking loss can impede feature learning.

\begin{table}[t]
    \centering
    \resizebox{0.425\textwidth}{!}{%
    \begin{tabular}{lcccc}
    \hline
     & \multicolumn{2}{c}{Market1501} & \multicolumn{2}{c}{PRID2011} \\ \cline{2-5} 
     & mAP & R-1 & mAP & R-1 \\ \hline
    w/o $\mathcal{L}^{\mathrm{m}}$ & 92.0 & 96.0 & 95.3 & 93.3 \\
    w/ $\mathcal{L}^{\mathrm{m}}$, $\lambda_2 = 1.0$ & \textbf{92.0} & \textbf{96.7} & \textbf{95.6} & \textbf{93.3} \\
    w/ $\mathcal{L}^{\mathrm{m}}$, $\lambda_2 = 2.0$ & 91.7 & 96.3 & 94.9 & 91.0 \\ \hline
    % \textcolor[HTML]{808080}{$\lambda_1 = \lambda_2 = 1.0$} & & & & \\
    w/ $\mathcal{L}^{\mathrm{r}}$, $\lambda_3 = 0.1$ & 92.0 & 96.6 & 96.3 & 94.4 \\
    w/ $\mathcal{L}^{\mathrm{r}}$, $\lambda_3 = 1.0$ & 91.8 & 96.8 & 97.3 & 95.5 \\
    w/ $\mathcal{L}^{\mathrm{r}}$, $\lambda_3 = 3.0$ & \textbf{92.2} & \textbf{96.9} & \textbf{97.3} & \textbf{95.5} \\
    w/ $\mathcal{L}^{\mathrm{r}}$, $\lambda_3 = 5.0$ & 92.0 & 96.6 & 96.0 & 93.3 \\
    w/ $\mathcal{L}^{\mathrm{r}}$, $\lambda_3 = 10.0$ & 91.1 & 96.6 & 95.4 & 93.3 \\ \hline
    w/ $\mathcal{L}^{\mathrm{a}}$, $\lambda_4 = 0.5$ & 91.9 & 96.7 & 94.2 & 93.3 \\
    w/ $\mathcal{L}^{\mathrm{a}}$, $\lambda_4 = 1.0$ & 92.0 & 96.8 & 95.0 & 92.1 \\
    w/ $\mathcal{L}^{\mathrm{a}}$, $\lambda_4 = 2.0$ & \textbf{92.0} & \textbf{97.0} & \textbf{98.1} & \textbf{97.8} \\
    w/ $\mathcal{L}^{\mathrm{a}}$, $\lambda_4 = 5.0$ & 91.1 & 96.5 & 94.7 & 92.1 \\
    w/ $\mathcal{L}^{\mathrm{a}}$, $\lambda_4 = 10.0$ & 89.7 & 96.2 & 93.9 & 92.1 \\ \hline
    \end{tabular}
    }
    \caption{Ablation studies of losses and balancing parameters in the unified feature learning and adaptation module.}
    \label{tab:bal}
\end{table}

Group 2 discusses the ratio of the regularization loss to feature and masking losses by keeping $\lambda_1$ and $\lambda_2$ fixed while modifying $\lambda_3$. Performance on both ReID tasks initially improves as $\lambda_3$ increases, peaking at 3.0 before decreasing. This reveals a trade-off: insufficient regularization leads to less robust and generalizable features, while excessive regularization results in overly smooth features that lack task-specific semantics.

Group 3 explores the impact of varying the ratio of alignment loss from video to frame. As this ratio increases, performance on both ReID tasks initially improves, peaking at 2.0 before decreasing. This trend demonstrates a balance between leveraging information from both modalities: insufficient alignment fails to adequately leverage video information, while excessive alignment overshadows frame information.

In summary, our experiments reveal the effectiveness of these losses. VILLS achieves best performance with $\lambda_1 = 1.0$, $\lambda_2 = 1.0$, $\lambda_3 = 3.0$, and $\lambda_4 = 2.0$. These values strike a balance between the different loss terms, enabling effective feature learning across both image and video modalities. We adopt these parameters in our final configuration.

\begin{figure}[t]
    \centering
    \includegraphics[width=\linewidth]{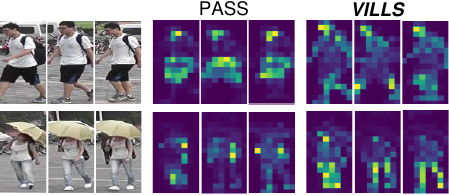}
    \caption{Visualization of attentions. VILLS demonstrates semantically consistent attention patterns. Please refer to the Supplemental Material for more details.}
    \label{fig:visual}
\end{figure}

\section{Conclusion}
This paper introduced VILLS, a novel self-supervised method that unifies image and video person re-identification within a single framework. VILLS addresses key limitations of existing modality-specific methods through two main components: 1) A Local Semantic Extraction module that adaptively extracts semantically consistent  features, and 2) A Unified Feature Learning and Adaptation module that captures temporal information and bridges the gap between images and videos while leveraging self-supervised learning on large-scale unlabeled datasets. These two modules allow VILLS to leverage complementary information from both domains, resulting in more robust and generalizable representations for ReID tasks. Our comprehensive experiments spanning image-based, video-based, and image-video mix ReID tasks, demonstrate VILLS's superior performance and generalization capabilities. VILLS consistently achieves state-of-the-art results, with significant improvements over existing methods in key metrics.

\section{Discussion of Potential Negative Societal Impact}
The authors affirm that all datasets utilized in this paper originate from public sources or have been approved by the subjects themselves. This research adheres to ethical guidelines and does not raise privacy or safety concerns. The objective of this paper is to enhance advancements in smart city applications and autonomous driving technologies.

%%%%%%%%% REFERENCES
{\small
\bibliographystyle{ieee_fullname}
\bibliography{egbib}
}

\end{document}

\title{Supplemental Material of VILLS \includegraphics[width=0.4cm]{1648590.png}: Video-Image Learning to Learn Semantics for Person Re-Identification}

\author{Siyuan Huang, Ram Prabhakar, Yuxiang Guo, Rama Chellappa, Cheng Peng\\
Johns Hopkins University\\
{\tt\small \{shuan124, rprabha3, yguo87, rchella4, cpeng26\}@jhu.edu}
}
\maketitle

\section{Fine-tuning and Inference}
For downstream ReID tasks, we use only the teacher's shared encoder and the resampler, discarding the student, all heads in the teacher, and the local semantic extraction module. We then fine-tune the teacher using ReID task losses, including cross-entropy loss for classifying different identities and triplet loss \cite{hermans2017defense} for clustering the same identity.

During inference, VILLS automatically extracts all features. For an input image, VILLS extracts coarse-grained and semantically consistent features. For an input video, VILLS also extracts temporal features. These features are then used for person matching, retrieval, or visualization based on the specific task.

\section{Implementation Details}
For the Unified Feature Learning and Adaptation module, we use a Vision Transformer (ViT) \cite{dosovitskiy2020image} as the shared encoder backbone. The resampler is a Perceiver Transformer \cite{jaegle2021perceiver}, which consists of a small transformer layer with cross attention. All heads are constructed using Multi-Layer Perceptrons with Batch Normalization \cite{ioffe2015batch}. For the local semantic extraction (LSE) module, we use a pre-trained Mask R-CNN \cite{he2017mask} from the COCO dataset \cite{lin2014microsoft}, while the interactive segmentation model is based on the pre-trained Segment Anything Model \cite{kirillov2023segment}. 

During pre-training, we train ViT-S and ViT-B on 8$\times$ A40 GPUs for 100 epochs. The training process takes approximately 100 and 200 hours for ViT-S and ViT-B, respectively. We use video batch sizes of 32 (ViT-S) and 16 (ViT-B), and image batch sizes of 128 (ViT-S) and 64 (ViT-B).  The video frame size is set to 64 $\times$ 44, with each video randomly sampling 8 frames. The image size is 256 $\times$ 128. The LSE-derived feature size is 128 $\times$ 64. The balancing parameters $\lambda_1$, $\lambda_2$, $\lambda_3$, and $\lambda_4$ are set to 1.0, 1.0, 3.0, and 2.0, respectively.

For downstream ReID tasks, we follow standard settings for each task and dataset. In PRCC and LTCC, the input size is 384 $\times$ 192. In Market1501, the input size is 256 $\times$ 128. In PRID2011 and MARS, the input size is $8 \times 256 \times 128$. In BRIAR-2, BRIAR-3, and BRIAR-4, the input is 384 $\times$ 128 for images and $8 \times 384 \times 128$ for videos. Unless otherwise specififed, all main results are conducted using ViT-B, while ablation studies are conducted using ViT-S.

\begin{figure*}[t]
    \centering
    \includegraphics[width=\linewidth]{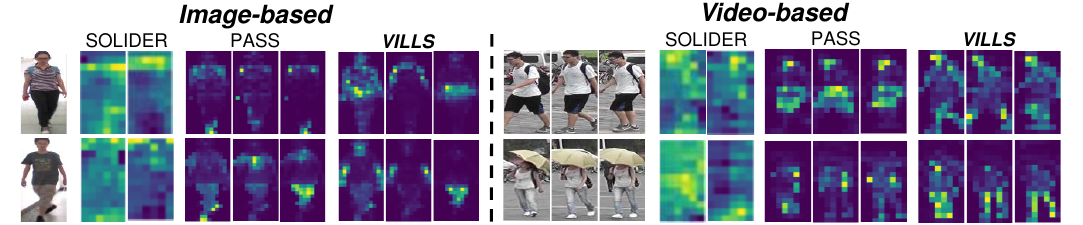}
    \caption{Visualization of attentions across different ReID methods. For SOLIDER, each attention map represents windowed attention. In image-based ReID, the first attention map for both PASS and VILLS is derived from coarse spatial features. The second and third attention maps for PASS are generated from local features, while for VILLS, they are generated from the LSE-derived features. In video-based ReID, all attention maps for PASS are derived from local features, whereas VILLS utilizes temporal features for all its attention maps. Notably, VILLS demonstrates semantically consistent attention patterns across both images and videos, highlighting its unified method to feature extraction.}
    \label{fig:visual}
\end{figure*}

\section{Visualization of Attention Maps}
Fig. \ref{fig:visual} compares attention maps between our method and others across different tasks. For image-based ReID, our coarse-grained spatial features outperform others. While existing methods lack a complete semantic concept of identity and focuses on peripheral parts, the proposed method clearly captures a complete identity with accurate focus. Moreover, our attention is semantically consistent, showing clear focus on specific body parts (e.g., arms, thighs), unlike others which have noise from unrelated areas. These results demonstrate the effectiveness of our method, particularly the LSE module, in extracting semantically consistent spatial features.

In video-based ReID, other methods lack temporal features, resulting in incomplete attention that fails to connect across frames. In contrast, the attention behavior in VILLS shows clear motion patterns highly consistent with the original video. Furthermore, the attention consistently focuses on the most significant motion parts of the identity, unlike other methods. These results highlight the effectiveness of UFLA module in successfully extracting temporal features. In summary, these visualizations showcase VILLS's ability to effectively and seamlessly extract both spatial and temporal features. These visualizations provide strong qualitative evidence for the effectiveness of VILLS in capturing semantically consistent and modality-appropriate features across various ReID tasks.

\section{Test Accuracy Curves}

\begin{figure}[t]
    \centering
    \includegraphics[width=\linewidth]{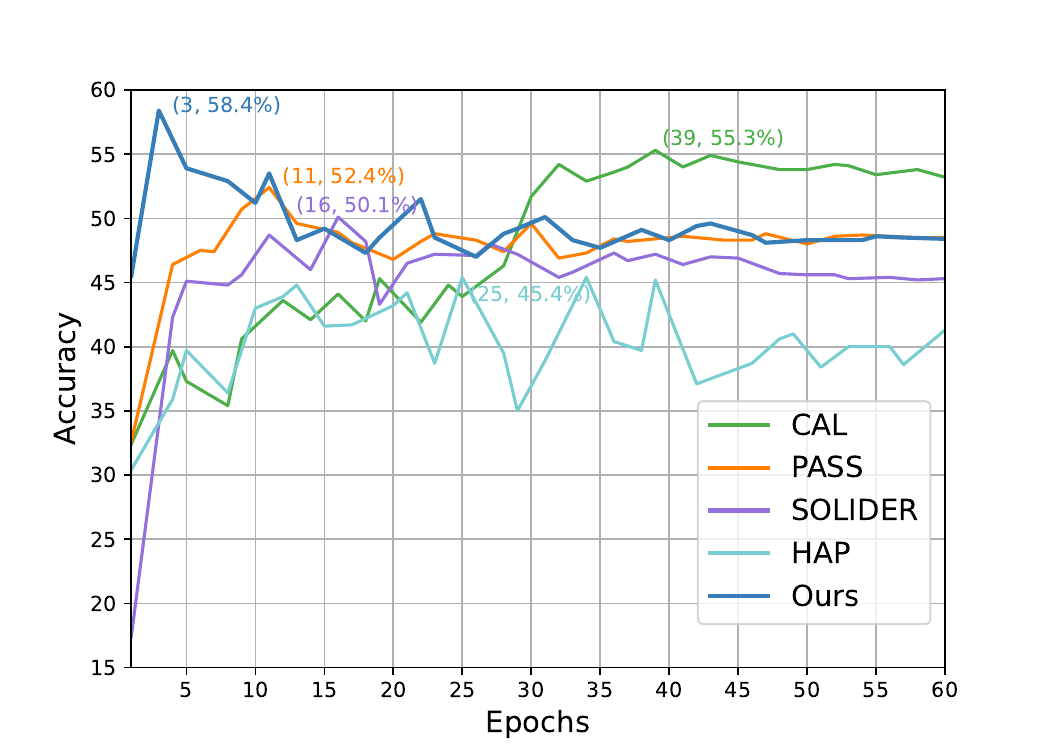}
    \caption{Test accuracy curves for various methods on the PRCC dataset. VILLS achieves the highest performance in the fewest epochs. This experiment was conducted using the image-only version of VILLS with ViT-B.}
    \label{fig:rank1}
\end{figure}

Fig. \ref{fig:rank1} illustrates the test accuracy curves for our method and state-of-the-art (SOTA) methods on image-based ReID. Notably, our method achieves a rank-1 accuracy of 58.4\% by epoch 3, outperforming other methods. In comparison, PASS (Zhu et al. 2022) reaches a rank-1 accuracy of 52.4\% in epoch 11, while CAL \cite{gu2022clothes} achieves 55.3\% in epoch 39, demonstrating the effectiveness of our method.

A comparison between pre-training methods (PASS \cite{zhu2022pass}, SOLIDER \cite{chen2023beyond}, HAP \cite{yuan2024hap}, and VILLS) and methods specifically designed for this dataset (e.g., CAL \cite{gu2022clothes}) reveals that pre-training methods converge faster. However, their performance  falls short of dataset-specific methods. VILLS stands out by not only surpassing other pre-training methods but also outperforming dataset-specific methods. Moreover, VILLS converges in fewer epochs, highlighting both its effectiveness and efficiency.

%%%%%%%%% REFERENCES
{\small
\bibliographystyle{ieee_fullname}
\bibliography{egbib}
}